\documentclass{article}

\usepackage[preprint]{neurips_2024}

\usepackage[pagebackref=true,breaklinks=true,colorlinks]{hyperref} 
\usepackage[utf8]{inputenc} 
\usepackage[T1]{fontenc}    
\usepackage{hyperref}       
\usepackage{url}            
\usepackage{booktabs}       
\usepackage{amsfonts}       
\usepackage{nicefrac}       
\usepackage{microtype}      
\usepackage{xcolor}         
\usepackage{subfigure}
\usepackage{graphicx}
\usepackage{amsmath}
\usepackage{amssymb}
\usepackage{times}
\usepackage{epsfig}
\usepackage{float}
\usepackage{placeins}
\usepackage{color, colortbl}
\usepackage{stfloats}
\usepackage{multirow}
\usepackage{xspace}
\usepackage{booktabs}
\usepackage{caption}
\usepackage{subcaption}
\usepackage{bbding}
\usepackage{enumitem}
\usepackage{tabularx}
\usepackage{dsfont}
\usepackage[hang,flushmargin]{footmisc}
\usepackage{mathtools}
\usepackage{amsthm}
\usepackage{pdfpages} 
\usepackage{wrapfig}
\usepackage{pifont}
\usepackage[capitalize,noabbrev]{cleveref}

\theoremstyle{plain}

\theoremstyle{definition}

\theoremstyle{remark}

\definecolor{blackpink}{rgb}{0.83, 0.19, 0.79}
\definecolor{darkgreen}{rgb}{0.00, 0.50, 0.10}
\definecolor{blue}{rgb}{0.00, 0.00, 1.00}

\def\sexyname{CoNav\xspace}
\def\unsure{\textcolor{black}}

\def\fjg{\textcolor{black}}

\definecolor{ballblue}{rgb}{0.13, 0.67, 0.8}

\usepackage[textsize=tiny]{todonotes}

\def\eg{\emph{e.g.,~}} 
\def\ie{\emph{i.e.,~}}

\def\mL{{\mathcal L}}

\def\0{{\bf 0}}
\def\1{{\bf 1}}

\newcommand\invisiblesection[1]{\noindent\textbf{#1}.}

\newcommand{\with}{\textcolor{darkgreen}{\ding{52}}}
\newcommand{\without}{\textcolor{red}{\ding{56}}}

\title{CoNav: A Benchmark for Human-Centered Collaborative Navigation}

\author{
    Changhao Li\textsuperscript{\rm 1,2}\thanks{Equal contribution.}~~~~
    Xinyu Sun\textsuperscript{\rm 1}\footnotemark[1]~~~~
    Peihao Chen\textsuperscript{\rm 1}\footnotemark[1]~~~~
    Jugang Fan\textsuperscript{\rm 1}~~
    Zixu Wang\textsuperscript{\rm 1}~~ \\
    \textbf{Yanxia Liu}\textsuperscript{\rm 1}~~
    \textbf{Jinhui Zhu}\textsuperscript{\rm 1}~~
    \textbf{Chuang Gan}\textsuperscript{\rm 3,4}~~
    \textbf{Mingkui Tan}\textsuperscript{\rm 1,5}\thanks{Corresponding author.} \\
    \textsuperscript{\rm 1}South China University of Technology,
    \textsuperscript{\rm 2}Pazhou Laboratory, \\
    \textsuperscript{\rm 3}UMass Amherst, 
     \textsuperscript{\rm 4}MIT-IBM Watson AI Lab,\\
    \textsuperscript{\rm 5}Key Laboratory of Big Data and Intelligent Robot, Ministry of Education \\
    changhaoli65@gmail.com
}

\begin{document}

\maketitle
\vspace{-2mm}
\begin{abstract}
Human-robot collaboration, in which the robot intelligently assists the human with the upcoming task, is an appealing objective. To achieve this goal, the agent needs to be equipped with a fundamental collaborative navigation ability, where the agent should reason human intention by observing human activities and then navigate to the human's intended destination in advance of the human. However, this vital ability has not been well studied in previous literature. To fill this gap, we propose a collaborative navigation (\sexyname) benchmark. Our \sexyname tackles the critical challenge of constructing a 3D navigation environment with realistic and diverse human activities. To achieve this, we design a novel LLM-based humanoid animation generation framework, which is conditioned on both text descriptions and environmental context. The generated humanoid trajectory obeys the environmental context and can be easily integrated into popular simulators. 
We empirically find that the existing navigation methods struggle in \sexyname task since they neglect the perception of human intention.
To solve this problem, we propose an intention-aware agent for reasoning both long-term and short-term human intention. The agent predicts navigation action based on the predicted intention and panoramic observation. The emergent agent behavior including observing humans, avoiding human collision, and navigation reveals the efficiency of the proposed datasets and agents. Our code will be released at \href{https://github.com/Li-ChangHao/CoNav.git}{https://github.com/Li-ChangHao/CoNav.git}.
  
\end{abstract}
\vspace{-1mm}
\section{Introduction}
\vspace{-1mm}
We study human-centered collaborative navigation for indoor environments as the first step toward efficient human-robot collaboration. 
In this task, the agent should act like a companion in perfect harmony with humans, and deduce human intention without being directly instructed. It actively predicts and navigates to the goal position where the human intends to go, ensuring the robot is ready to provide further assistance. This collaborative navigation skill is vital in various application scenarios, including household robots, healthcare assistance, and elder care. 
%

Learning a collaborative navigation agent requires a human-centered navigation dataset that incorporates humanoids interacting with the environment. 
However, constructing such a dataset is challenging for two reasons:
1) it is hard to integrate meaningful and realistic human motions conditioned on the 3D environment; 2) collecting animation of different human activities is expensive and thus the dataset is difficult to scale.
%

Towards the collaborative navigation goal, prior methods~\cite{chen2019crowd,liu2020decentralized,liu2022intention,iGibson} introduce a social navigation task, focusing on minimizing human-robot collisions. In these works, some rigid human bodies wander aimlessly around the environment without any body movements. Due to the lack of semantic relations between human trajectory and environmental context, realistic human behavior and reaction can not be well represented. More recent work~\cite{habitat3.0} introduces human activities animation by rendering and replaying pre-recorded human poses. Although it provides a strong interface to control human avatars, the rendered human animations are still limited by the diversity of pre-recorded poses and have no semantic relation with environmental context.
More importantly, most of the existing works failed to consider how an intelligent robot collaborates with a human beyond simply keeping a distance from it. 


\begin{figure}[!t]
    \centering
    \includegraphics[width=\linewidth]{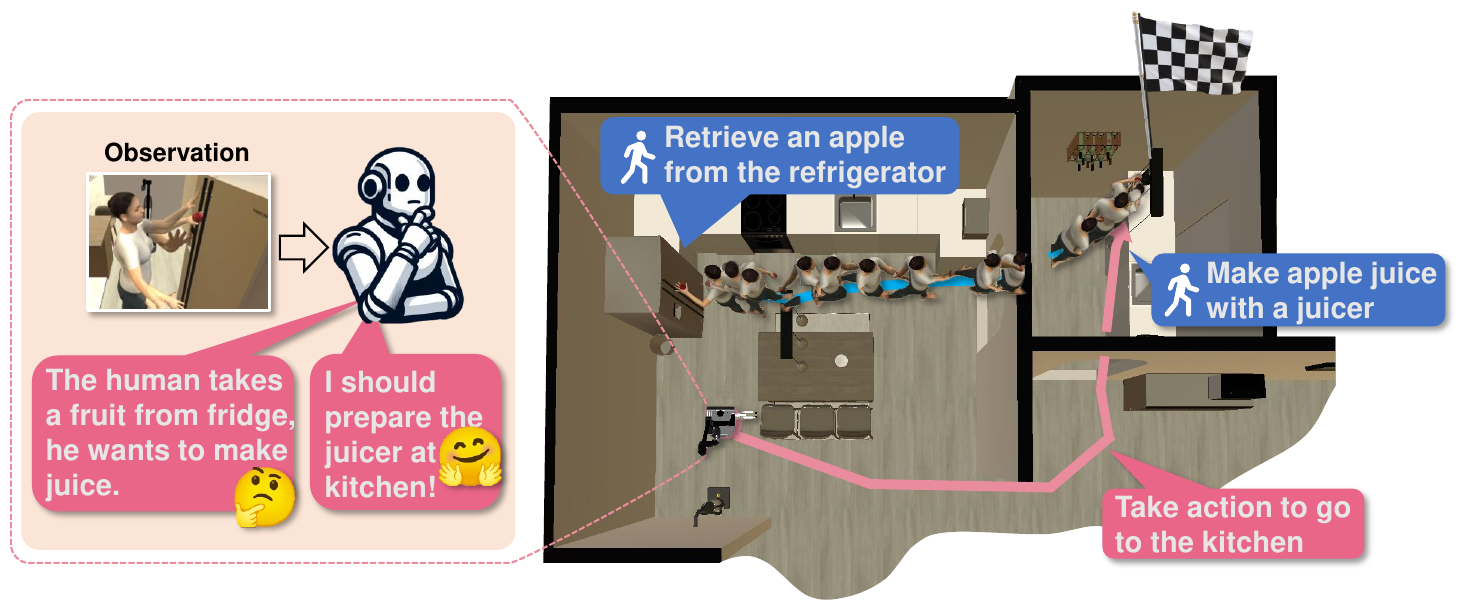}
    \caption{Illustration of an example episode in our \sexyname task. The agent observes the human activity and then predicts and navigates to the human's intended destination.
    }
    \label{fig:teaser}
    \vspace{-8mm}
\end{figure}

In this work, we aim to construct a new navigation benchmark, named collaborative navigation (CoNav), with diverse and realistic humanoid animations that reveal daily human activities in the indoor scenario. The agent observes these activities, reasons the human intended destination, and navigates to the destination in advance of the human. The CoNav task illustration is shown in Figure~\ref{fig:teaser}.

To generate diverse and realistic humanoid animation that obeys environmental context, we take advantage of the reasoning ability of Large Language Models (LLMs)~\cite{openai2023gpt4, chiang2023vicuna, ouyang2022training} open-vocabulary motion generation ability of Generation Models (GMs)~\cite{PriorMDM, jiang2023motiongpt, MDM, chen2023executing}.
Our LLM-based controllable animation generation framework consists of three steps: 1) environment-aligned activities reasoning; 2) open-vocabulary action animation; and 3) environment-aligned trajectory incorporation. Specifically, we first describe all environmental objects to an LLM and prompt it to generate reasonable activities conducted with the given objects. The activities follow the template  ``\texttt{do [activity\_A] with \{object\_A\}, then go to another place and do [activity\_B] with \{object\_B\}}". A chain-of-though technique is applied with few-shot examples to make sure two activities link up causally and have semantic correlations with the objects. 
Then, the activity descriptions are fed into the text-conditioned motion generation models~\cite{jiang2023motiongpt} to obtain humanoid animations. 
Since each activity is associated with an object in the environment, we incorporate the generated humanoid animation in front of the corresponding objects. To connect the two activities, we plan a path from object A to object B and utilize a trajectory-controllable motion generation method~\cite{PriorMDM} to generate a walking animation following the planned path. We get an environment-aligned trajectory by combining two activity animations and a walking animation. In total, we automatically generate more than 25k diverse and realistic trajectories in 49 environments. A comparison with the existing datasets is shown in Table~\ref{benchmarks-compare}. We are the first to combine a text-motion model with the simulator to generate diverse humanoid activities in 3D simulation environment.

We further develop an intention-aware agent for CoNav task. The agent consists of two components: a human intention predictor; and a policy for predicting navigation action based on the predicted human intention. Specifically, we represent human intention from both long-term and short-term perspectives. The long-term intention includes the next activity and the next object to be interacted with, while the short-term intention includes the predicted trajectory for the next few steps. We leverage a Perceiver model~\cite{Perceiver} to predict the long-term intention and use a trajectory prediction model~\cite{gst} to predict the short-term intention. The policy takes as the input of the panoramic RGB-D observation and the predicted intention to determine the next navigation action. We train the intention predictor and policy using supervised learning and imitation learning, respectively. 
The trained agents perform the emergent behavior of observing human activities, avoiding collision with humans, and navigating to the destination that the human intent to go. This reveals that \sexyname is a superior benchmark for building human-centered collaborative agents to assist humans in everyday scenarios.


Our main contributions are three folds:
1) We propose a novel collaborative navigation task, \sexyname, which is challenging and cannot be easily tackled with existing methods; 2) a human-centered navigation dataset with realistic and diverse human animations is proposed for developing agent with CoNav ability; 3) We empower an agent with intention perception skills and it yields promising performance on the new proposed CoNav task.

\begin{table}[t!]
\small
\centering
\begin{tabular}{@{}lllllll@{}}
\toprule
Method     & \begin{tabular}[c]{@{}l@{}}Dynamics\\ Model\end{tabular} & \begin{tabular}[c]{@{}l@{}}Navigation\\ Settings\end{tabular}  & \begin{tabular}[c]{@{}l@{}}Scene-human\\ Interaction\end{tabular} & \begin{tabular}[c]{@{}l@{}}Coherent\\ Motion\end{tabular} & \begin{tabular}[c]{@{}l@{}}Diverse\\ Activities\end{tabular} & \begin{tabular}[c]{@{}l@{}}Relistic\\ Walking\end{tabular} \\ \midrule
iGibson-SN~\cite{iGibson}            & ORCA  & Point-goal  & \without & \without & \without & \without \\
HabiCrowd~\cite{vuong2023habicrowd}  & UPL++ & Object-goal & \without & \without & \without & \without \\
Habitat3.0~\cite{habitat3.0}         & ORCA  & Social-goal & \with    & \with    & \without & \without \\ \midrule
\sexyname                      & Prior-MDM & Intended-goal & \with    & \with    & \with    & \with    \\ \bottomrule
\end{tabular}
\caption{We compare our \sexyname with existing human-centered navigation benchmarks from different perspectives. We are the first work to incorporate a text-driven dynamic model into the simulator and thus yield high-fidelity human animations with diverse activities, which also enables interaction with scene benefit from our \unsure{environment conditioned motion generation framework}.}
\label{benchmarks-compare}
\vspace{-8mm}
\end{table}

\vspace{-1mm}
\section{Related Work}
\vspace{-1mm}
\label{sec:related}

\subsection{Indoor Navigation}
Indoor Navigation is a fundamental and crucial task for robots to perform various downstream household tasks~\cite{iGibson, Batra2020rearrange, ai2thorRearrange, BEHAVIOR, BEHAVIOR-1K}. In indoor navigation tasks, the agent receives a geometry~\cite{habitat19iccv} or semantic~\cite{Dhruv2020ObjectNav, zhu2017ImageNav, sun2024prioritized} goal, and then makes efforts to navigate to this goal. According to the goal type, indoor navigation can be divided into different categories: point-goal navigation~\cite{habitat19iccv,DDPPO}, object-goal navigation~\cite{SemExp,OVRL-V2,ZSON}, image-goal navigation~\cite{OVRL,OVRL-V2,ZER,fgprompt}, vision-language navigation~\cite{R2R, RxR, LookBeforeYouLeap, LM-Nav, A2NAV}, audio-visual navigation~\cite{gan2019look, chen2020soundspaces, Move2Hear}, etc. Taking image-goal navigation~\cite{fgprompt} as an example, the agent needs to explore the environment, locate the target object specified by the image, and navigate to its proximity. In this work, we focus on a more challenging task of indoor navigation, human-centered navigation, in which the agent is required not only to navigate to the goal effectively and efficiently but also to cooperate with moving humans in the surrounding environment. \fjg{Different from existing benchmarks~\cite{iGibson, Nathan2022SEAN, habitat3.0} towards human-centered navigation which focuses on agents and humans in isolation, our proposed benchmark endows humans with a wide variety of activities, additionally demands that agents observe human actions to predict their intentions and navigate to potential destinations ahead of humans.}


\subsection{Human Motion Generation}
An environment occupied by moving humans is essential for human-centered navigation, however, most existing datasets have human only~\cite{Mvp-Human, RenderPeople, Dynacam, Yi2023Generating} or 3D scenes only~\cite{3D-FRONT, MP3D,HM3D,ScanNet,Replica}. Recently, some appealing works researched to generate scenes given 3D human motion~\cite{MIME} or generate human poses given a 3D scene~\cite{Yan2020Generating, Minseok2023Pose}. ~\cite{MIME} collect a dataset by adding contact humans from RenderPeople~\cite{RenderPeople} into the 3D scene from 3D-FRONT~\cite{3D-FRONT}, and randomly assigning plausible interactions between humans and objects, which has only three types of interactions and is less scene-related.

To generate more scene-constrained and long-horizon human motion, recent methods use large models
~\cite{CLIP, Imagen, DDPM} that have pretrained on Internet-scale text-images pairs and various motion transition methods. ~\cite{MotionCLIP} use well-aligned text-image embedding space to perform text-motion generation. 
~\cite{MDM} extend text-to-image diffusion models to execute text-to-motion tasks, and ~\cite{PriorMDM} further endow the capability of generating long-sequence motion through DoubleTake based on ~\cite{MDM}.
~\cite{Mir2023Generating} first infer interaction keypoints based on segmented scene pointcloud and language instructions and blend the long-sequence motion between action keypoints via dividing the whole motion into walk motion, transition-in motion, and transition-out motion. 
~\cite{Pi2023Hierarchical} leverage VQ-VAE~\cite{VQ-VAE, VQ-VAE-2} and transformer DDPM~\cite{DDPM} to generate goal pose and a number of milestones, and then infill motions between milestones to complete a full trajectory. 
~\cite{Huang2023Diffusion} propose SceneDiffuser that models the scene-constrained goal-oriented trajectory by conditioning the diffusion model on 3D scenes and guiding the model with physics-based objectives during inference time. 
\vspace{-1mm}
\section{\sexyname Benchmark}
\vspace{-1mm}
\label{sec:method}

\begin{figure*}[!t]
    \centering
    \includegraphics[width=1.0\linewidth]{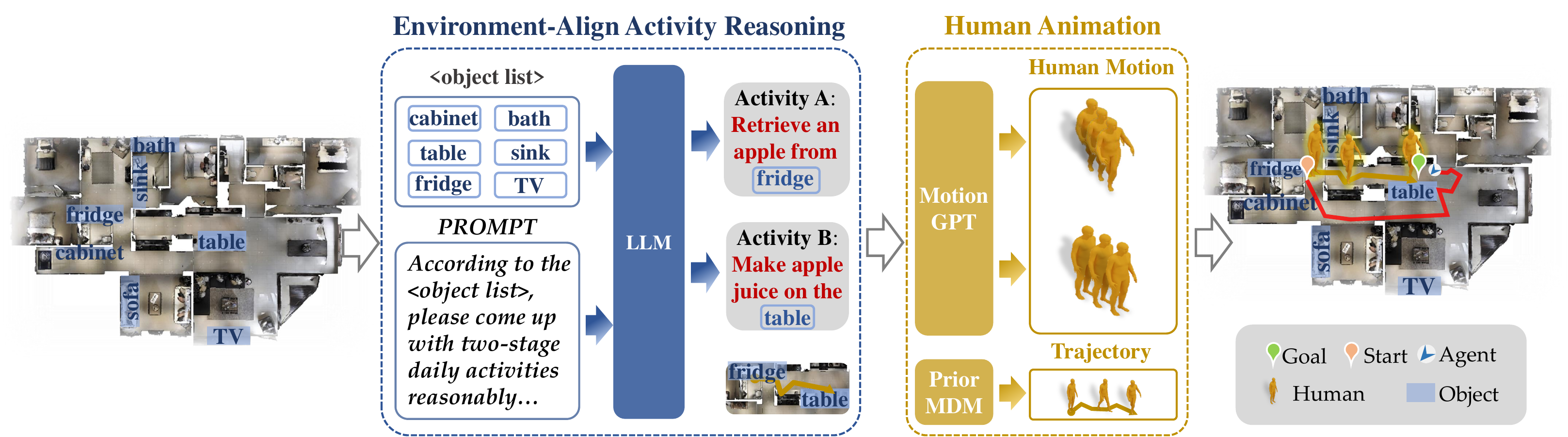}
    \caption{Overview of \sexyname dataset generation. We first use LLMs to reason environment-aligned causal activities. Then, generative models are used to animate activity and walking motion.
    }
    \label{fig:overview}
    \vspace{-7mm}
\end{figure*}

We built the \sexyname benchmark to assess the agent's intention prediction ability through human behavior understanding and common sense reasoning for efficient human-robot collaborative navigation in the indoor environment.
We first generate contextual action descriptions by bootstrapping the instruction-following ability of a Large Language Model~\cite{openai2023gpt4,ouyang2022training}, and then blend two motion generation models MotionGPT~\cite{jiang2023motiongpt} and PriorMDM~\cite{PriorMDM} to generate human motion based on text descriptions and given trajectories. Finally, we collect the navigation dataset using human action rendering API in Habitat 3.0~\cite{habitat3.0} simulator.
In the following paragraph, we first introduce the definition of CoNav task and then the data generation process.

\subsection{CoNav Task Definition}

In a \sexyname task, a human performs daily home routines consisting of two-stage activities in sequence, namely \texttt{[activity\_A]} and \texttt{[activity\_B]}. As shown in Figure~\ref{fig:teaser}, these two activities are causally related, following the logic of  ``in daily life, humans usually do activity B after finishing activity A''. The agent should actively observe human movement, figure out the intended destination of the human for doing activity B, and navigate to that target location before the human finishes activity B. Additionally, following previous social navigation literature~\cite{habitat3.0}, the agent must avoid collision with humans when navigating.


\noindent\textbf{Observation and action spaces.}
\label{sec:obs-space}
The agent is equipped with a head-mounted panoramic RGB-D camera and a GPS \& compass sensor that informs the current agent pose $P^a$. We also provide the agent with the human position $P^h$, which helps the agent to locate and observe the human movement. The agent has a continuous action space, including linear velocity $V_l$ and angular velocity $V_a$ with a maximum speed of 40 meters/s and 30 rad/s, respectively.

\noindent\textbf{Success criterion.}
We define a definite success criterion where navigation is considered successful only when the agent reaches the position of second-stage activity within 1m without colliding with the person. Besides, an episode will be judged as a failure if the agent fails to navigate to the target location in time after the human completes the second-stage activities. 

\subsection{CoNav Dataset Generation}
The CoNav task includes human animations of diverse causally related activities in environments. The existing datasets only contain rigid human bodies or random wandering animation, which can not meet the requirements of the CoNav task. In this paper, we first leverage LLMs to reason causal activities that are executable in the given environment. Then we animate the motion of these activities and human walking using motion generation models. Last, we incorporate the generated animation into an appropriate position of environment. The data generation pipeline is shown in Figure~\ref{fig:overview}.

\noindent\textbf{Environment-aligned causal activities reasoning by LLMs.}
We collect the CoNav dataset upon the scenes in HSSD dataset~\cite{HSSD} with object coordinate and category annotation. We describe all these objects to an LLM and use a chain of thought to prompt it to generate two-stage activities that can be executed using these objects. We prompt LLMs to ensure that each activity is associated with one object and give the reason why these two activities are causal. More details about the chain-of-though prompt design can be seen in the Appendix\ref{supp:data-generation}. 
Since the generated activity descriptions are in open-vocabulary and may have noise, we conduct further filtering steps including 1) excluding texts whose objects were not in the environment and 2) computing BERT embedding similarity between different activity descriptions and removing those with high similarity. 

\noindent\textbf{Motion fragment generation from text description.}
We aim to synthetic realistic and diverse human motion in a single fragment. With the state-of-the-art text-driven motion generation results achieved by the MotionGPT~\cite{jiang2023motiongpt}, we apply it to each of our generated activity descriptions to obtain the motion fragment in the form of Skinned Multi-Person Linear model~\cite{SMPL} (SMPL). Note that the motion fragment in SMPL format is further transformed into the HumanML3D~\cite{HumanML3D} style for the subsequent motion fragments completion procedure. As a result, for each motion description, we acquire a human motion fragment denoted by $p^{1:T} = (r^{1:T}, j^{1:T})$, where $T$ is fragment length, $r$ is the root parameter (including the angular velocity along Y-axis, the linear velocities on XZ-plane and the height of the root) and $j$ is the joint parameter (consisting of the local joints position, velocities and rotations in root space). At last, we generate motion fragments $p^{1:T_A}_A$ and $p^{1:T_B}_B$ for \texttt{ [activity\_A]} and \texttt{[activity\_B]}, respectively. A visualization example of the generated activity is shown in Figure~\ref{fig:motion} (a).


\noindent\textbf{Motion fragments completion.}
With the motion fragments $p^{1:T_A}_A$ and $p^{1:T_B}_B$ respectively from the source and target locations, we need to ``fill the gap'' between them to acquire a complete human motion sequence. To this end, we turn to a state-of-the-art motion diffusion model, PriorMDM~\cite{PriorMDM}, to complete the missing motions by generating a \textit{walking motion fragment}. To be specific, we initialize the walking fragments as $p_W^{1:T_W} = (r_W^{1:T_W}, j_W^{1:T_W})$, where the root parameters $r_W^{1:T_W}$ are transformed from the shortest path from $A$ to $B$ and $j_W^{1:T_W}$ are draw from the standard Gaussian distribution, \eg~$j_W^t \sim \mathcal{N}(\0, \1), t=\{1,\dots,T_W\}$. Then, we concatenate $p^{1:T_A}_A$, $p_W^{1:T_W}$, $p^{1:T_B}_B$ in the time dimension for the input of PriorMDM and get the generated joint parameters of the walking segments \eg~$j_W^{1:T_W}$.\footnote{Limited by the context length of PriorMDM, we truncate part of the data located at the head of $p^{1:T_A}_A$ and the tail of $p^{1:T_B}_B$.} Here, we abuse the notation of the randomly initialized and generated joint parameters for simplicity. An example of the generated trajectory is shown in Figure~\ref{fig:motion} (b).

\noindent\textbf{Injecting human motion into simulator.}
Following popular benchmarks such as Habita3.0~\cite{habitat3.0} that controls human action with SMPL-X parameters, we first transform our completed motion fragments from the format of HumanML3D into the SMPL-X. More importantly, we make two further refinements to seamlessly blend human action into the environment. \textbf{First}, we adjust the humanoid global posture towards the target object. \textbf{Second}, we make the humanoid pickup the target object. In this sense, the humanoid presents a natural interaction with both the environment and the related objects. Note that incorporating human action into the environment is very efficient thanks to the Habitat 3.0 implementation.


\vspace{-1mm}
\section{Intention-Aware Agent for CoNav Task}
\begin{figure}[t!]
  \centering
  \begin{minipage}[t]{0.4\textwidth}
    \centering
    \vspace{0pt}
    \includegraphics[width=\linewidth]{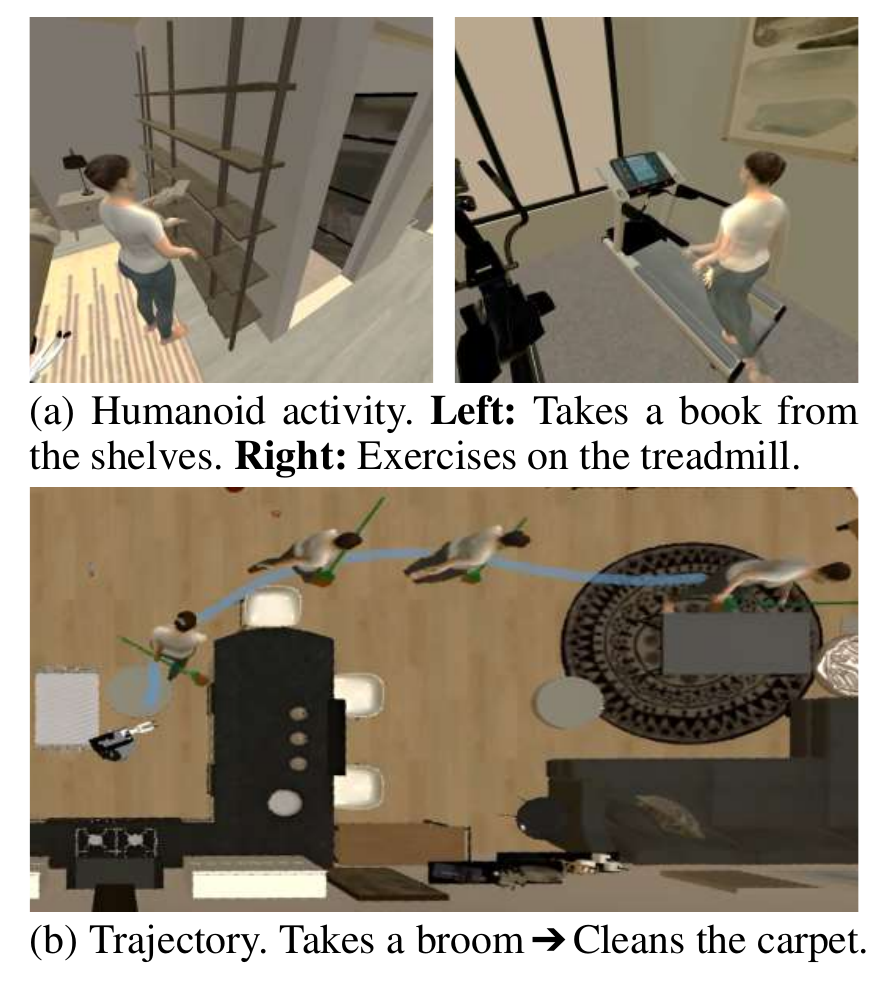}
    \vspace{-6.5mm}
    \caption{Visualization of generated (a) humanoid activity and (b) trajectory.}
    \label{fig:motion}
  \end{minipage}
  \hfill
  \begin{minipage}[t]{0.56\textwidth}
    \centering
    \vspace{0pt}
    \includegraphics[width=\linewidth]{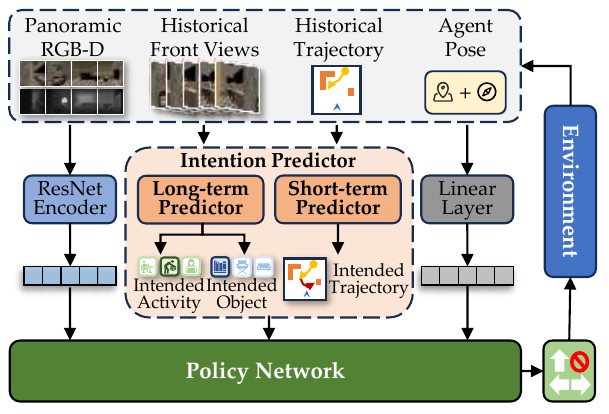}
    \caption{Overall architecture of our intention-aware agent. The policy takes as input the predicted intention, the panoramic RGB-D, and the agent pose for determining navigation actions.}
    \label{fig:agent}
  \end{minipage}
  \vspace{-4mm}
\end{figure}

\vspace{-1mm}
We target building an intelligent agent that can understand human behavior and then take navigation actions to reach a location suitable for assisting the human. To this end, we equip the agent with an intention predictor for reasoning humans' long-term intention, \ie the next activity and the next interacted object, and short-term intention, \ie where the human would go in the next few steps. The agent takes the predicted intention, panoramic observations, and pose as input to determine navigation actions. The paradigm of the proposed intention-aware agent is shown in Figure~\ref{fig:agent}.

\subsection{Long-Shot-Term Intention Predictor}
The activities we perform are causally related in most cases. Performing one action can directly lead to another action involving a specific object. For example, when we observe a person retrieving an apple from the refrigerator, it is reasonable to guess that they might go to find a juicer to make apple juice. Motivated by this fact, we consider the intended activity and the interacted object as representations of humans' long-term intentions. However, in some cases, performing one activity may lead to multiple possible subsequent actions. After grabbing an apple, the person might also head to the sofa to sit and eat the apple. The agent should track the human's trajectory and predict their next few steps. We consider this short-term intention, which provides more concrete clues to the agent when determining navigation actions.



\noindent\textbf{Long-term intention predictor.}  
We predict the human's long-term intention by observing human's historical movement. Specifically, we take as input a sequence of front-view RGB images from the panoramic camera. These images are fed into a Perciver model~\cite{Perceiver} to extract the human movement features, followed by two classification heads for predicting intended object $\hat{Y}_{obj_\mathrm{B}}$ and intended activity $\hat{Y}_{act_\mathrm{B}}$, respectively. We chose the Perceiver model as our long-term intention predictor because it is compatible with inputs of varying frame numbers. Since the agent observes streaming images, at the beginning of an episode, it can simply feed the images collected up to the current step into the Perceiver model. During training, the long-term intention predictor takes as input a random number of frames about activity A and predicts activity B out of 107 possible activities and object B out of 51 possible objects. We use cross-entropy loss for training.

\noindent\textbf{Short-term intention predictor.}  
The short-term intention, \ie human's moving position in the subsequent few steps, is predicted by a pedestrian trajectory predictor GST~\cite{gst}. GST is a time series model that takes as input the historical human trajectory $T_{t-k:t} = \{ P^h_{t-k}, ..., P^h_{t-1}, P^h_{t} \}$ and recursively predicts the future trajectories $\hat{T}_{t+1:t+5} \{ \hat{P}^h_{t+1}, ..., \hat{P}^h_{t+4}, \hat{P}^h_{t+5} \}$. Since pedestrian trajectory prediction is not the focus of our work, we use a pre-trained GST model~\cite{gst} and freeze it when learning the CoNav agent.



\subsection{Learning Intention-Aware Agent}
Given the predicted long-term intention $\hat{Y}_{act_\mathrm{B}}$, $\hat{Y}_{obj_\mathrm{B}}$ and short-term intention $\hat{T}_{t+1:t+5}$, we feed them together with agent pose $P^a_t$, and panoramic RGB-D features to an LSTM policy for predicting the next action $\hat{V}_l$ and $\hat{V}_a$. The panoramic features are extracted by a ResNet-50. We train the CoNav agent using imitation learning by optimizing the following L2 loss: 
\begin{equation}
\begin{aligned}
    \mL  &= {(V_l-\hat{V}_l)^2} + {(V_a-\hat{V}_a)^2} \\
\end{aligned}
\end{equation}

Instead of directly going to the goal position, we design a heuristic strategy for collecting ground-truth actions $V_l$ and $V_a$ that allows the agent to first observe the human and then proceed to the human's intended destination. Specifically, the agent first locates and approaches the human, then stops for a fixed number of steps K to observe the human's movements, and finally moves to the goal position. Using the positions where the human performs activities A and B, we employ the A* algorithm~\cite{A*l} to plan the ground-truth path for the agent. Additionally, we enable the agent to avoid collisions with humans to adhere to the safety guidelines of social navigation. During training, we provide the agent with the ground truth of the human's intentions. 

During inference, the predicted long-term intentions change frequently as the agent receives different observations of the human. Sometimes,  the agent cannot see the human due to occlusion or moving direction. This leads to an inaccurate long-term intention prediction. To address this, we keep track of the historically predicted long-term intentions and select the top 5 predictions with the lowest entropy. We then average these 5 predictions to provide a more reliable input for the agent.

\vspace{-1mm}
\section{Experiments}
\vspace{-1mm}
\label{experiments}
Based on the generated CoNav dataset, we evaluate both the existing navigation methods and our proposed intention-aware agent for human-centered collaborative navigation task. We also evaluate the effectiveness of different types of predicted intention. 

\subsection{Baselines}
We establish several baseline methods and investigate how they perform in our proposed \sexyname task, including a heuristic pioneer agent, a pre-trained object-goal navigation agent, a state-of-the-art image-goal agent, and a human-following agent released by Habitat 3.0. Details are as follows:
\begin{itemize}[leftmargin=*]
\setlength{\itemsep}{2pt}
\item \textbf{Pioneer Agent:} This is a heuristic navigation agent. The agent is coded for finishing the \sexyname task by navigating to the location where the human is intended to go, and this is why it is called ``\textit{pioneer}''. To achieve this goal, we calculate the next waypoint the human will go based on its orientation and speed, and endow the agent with a ground-truth pathfinder which can directly translate the agent to the planned waypoint. 

\item \textbf{Object-goal Agent~\cite{homerobot,ZSON}:} Since the \sexyname task can be deemed as an object-goal navigation task that requires the agent to infer the secondary object based on the observations on the human, we try to tackle this task by utilizing recent success in object-goal navigation literature. For this baseline, we construct two different methods to infer the secondary object, one is querying a Large Vision Language Model~\cite{liu2023llava, zhu2023minigpt, openai2023gpt4} (LVLM) based on the first object category and visual observation input, and another variant directly input the oracle object category.

\item \textbf{Image-goal Agent~\cite{fgprompt,ZSON,ZER}:} Another popular serial of navigation approaches~\cite{fgprompt,ZSON,ZER} studies how to perform navigation given a goal image to determine the destination. We test the most well-performed image-goal navigation method~\cite{fgprompt} on our \sexyname task by providing the agent with a third-person RGB image that captures near the human at each time step. The agent is tasked to find and catch up with the humanoid avatar given the goal image. Since the agent is trained to stop within 1 meter of where the goal image was taken, the agent is expected to follow the human until it reaches the destination.

\item \textbf{Human-following Agent~\cite{habitat3.0}:} Another possible solution for our \sexyname task is walking closely with the human as previous endeavor done in Habitat 3.0~\cite{habitat3.0}. We directly test the agent released by~\cite{habitat3.0}, which is trained using RL with a human-following reward.

\end{itemize}

\subsection{Experimental Setup}
\invisiblesection{CoNav Dataset Statistic}
Follow the Habitat 3.0 scene split, we divide the HSSD scenes~\cite{habitat3.0} into 37 training scenes and 12 validation scene, generating 20,293 training episodes and 4,962 evaluation episodes for CoNav dataset. The training and validation scenarios do not overlap. In our dataset, we have a total of 51 types of objects and 107 types of activities. We visualize some generated trajectories and human motion in Figure~\ref{fig:motion}. Please refer to Appendix~\ref{supp:dataset-analysis} and Appendix~\ref{supp:humanoid-animation} for more data statistics and visualization.


\noindent\textbf{Agent Configuration.}
We set up a wheeled chassis agent with a height of 1.25m and, a radius of 0.3m.
For all experiments, we equip the agent with a panoramic RGB-D observation sized $112\times448\times4$ placed 1.25m from the ground, a robot GPS \& compass sensor, and additional human position observations mentioned in Section~\ref{sec:obs-space}. 
We train the agent with ground-truth shortest paths using imitation learning for 50 epochs. We use 8 NVIDIA GeForce RTX 3090 GPUs for training with a batch size of 512. It takes around 4 hours for training.
After training, we test the agent on the validation episodes. At the start of all validation episodes, the agent is placed near the humanoid to ensure it can observe the human activity.



\noindent\textbf{Evaluation metrics.} We use six metrics to evaluate agents on the \sexyname task. 
\begin{itemize}[leftmargin=*]
    \item {First-to-Arrive Success Rate} (FASR) is defined as the success ratio of the number of episodes. Once the agent reaches a location within 1 meter of \texttt{[object\_B]} before the human arrives, we consider this trail a success episode. This metric motivates the agent to locate its target through predictive analysis of human behavior, rather than just trailing the human. 
    \item {Robot-Arrive Success Rate} (RASR) covers the success similar to the FASR but allows the agent to arrive after the human. If the agent can reach within 1m of \texttt{[object\_B]} before the humanoid completes \texttt{[activity\_B]}, we consider it a success.
    \item {FASR Weighted by Path Length} (FASPL) is the weighted version of FASR metric to evaluate the navigation efficiency. Let $l$ be the minimum steps taken for the shortest path to the destination, and $p$ be the agent's path steps, $\text{FASPL}=\text{FASR}\times \frac{l}{\max(l,p)}$.
    \item {RASR Weighted by Path Length} (RASPL) is the weighted version of RASR metric to evaluate the navigation efficiency. The formula is similar to FASPL, where $\text{RASPL}=\text{RASR}\times \frac{l}{\max(l,p)}$.
    \item {Collision Rate} (CR) focuses on the frequency of human-robot collision. It measures an agent's ability to avoid collisions with humanoids.
    \item {Average Steps} (AS) indicates the average number of steps the agent takes for an episode.
\end{itemize}

\begin{table*}[t]
\small
\centering
\resizebox{\textwidth}{!}{
\begin{tabular}{lllllll}
\toprule
Methods     & \multicolumn{1}{c}{FASR↑} & \multicolumn{1}{c}{RASR↑} & \multicolumn{1}{c}{FASPL↑} & \multicolumn{1}{c}{RASPL↑} & \multicolumn{1}{c}{CR↓} & \multicolumn{1}{c}{AS↓}\\ 
\midrule
Random Agent & 4.5 & 4.6  & 3.7 & 3.7  & 32.7 & 442.6 \\ 
Pioneer Agent (Oracle Pathfinder) & 61.6 & 61.8  & 46.6 & 46.7  & 30.4  & 308.1   \\
\midrule
Human-following Agent{{}}~\cite{habitat3.0} & 3.5 & 3.7  & 1.6 & 1.7  & 49.0  & 350.8   \\ 
Object-goal Agent (Oracle Goal Object)$^\dagger$~\cite{ZSON}  & 15.6 & 16.4 & 10.7 & 10.9 & 46.1 & \textbf{295.5}  \\ 
Object-goal Agent (Predicted Goal Object)$^\ddagger$~\cite{ZSON}  & 8.6 & 9.3 & 7.2 & 7.5 & 38.7 & 377.4   \\ 
\unsure{Image-goal Agent}~\cite{fgprompt} & 7.6 & 8.1 & 6.0 & 6.2 & 41.4 & 372.5 \\ 
\midrule
\sexyname Agent (Ours) & \textbf{19.8} & \textbf{23.2} & \textbf{13.5} & \textbf{14.9} & \textbf{20.1} & 354.5  \\
\bottomrule
\end{tabular}}
\caption{Evaluation results of different baselines on the validation split of our \sexyname benchmark. $^\dagger$means oracle object categories are used. $^\ddagger$means we utilize an external LLM to deduce the goal object based on the sensory observations.}
\label{tab:sota}
\vspace{-2mm}
\end{table*}

\subsection{Result Analysis}
\label{sec:ablation}

\begin{table}[t]
\small
\centering
\label{tab:my-table}
\begin{tabular}{@{}cccccccccc@{}}
\toprule
\multirow{3}{*}{Exp. \#} & \multicolumn{2}{c}{Long-Term} & Short-Term & \multicolumn{1}{l}{\multirow{3}{*}{FASR↑}} & \multicolumn{1}{l}{\multirow{3}{*}{RASR↑}} & \multicolumn{1}{l}{\multirow{3}{*}{FASPL↑}} & \multicolumn{1}{l}{\multirow{3}{*}{RASRL↑}} & \multicolumn{1}{l}{\multirow{3}{*}{CR↓}} & \multicolumn{1}{l}{\multirow{3}{*}{AS↓}} \\ \cmidrule(lr){2-4}
  & Intended      & Intended      & Intended   & \multicolumn{1}{l}{} & \multicolumn{1}{l}{} & \multicolumn{1}{l}{} & \multicolumn{1}{l}{} & \multicolumn{1}{l}{} & \multicolumn{1}{l}{} \\
  & Activity      & Object        & Trajectory & \multicolumn{1}{l}{} & \multicolumn{1}{l}{} & \multicolumn{1}{l}{} & \multicolumn{1}{l}{} & \multicolumn{1}{l}{} & \multicolumn{1}{l}{} \\ \midrule
1 & \without   & \without   & \without   & 14.5 & 17.0 & 10.6 & 11.8 & \textbf{14.7} & 402.9 \\
2 & \with      & \without   & \without   & 14.4 & 17.4 & 10.2 & 11.9 & 27.7 & \textbf{330.1} \\
3 & \without   & \with      & \without   & 15.7 & 18.6 & 10.6 & 11.7 & 16.0 & 391.9 \\
4 & \with      & \with      & \without   & 17.5 & 20.4 & 11.4 & 12.8 & 22.2 & 352.5 \\
5 & \with      & \with      & \with      & \textbf{19.8} & \textbf{23.2} & \textbf{13.5} & \textbf{14.9} & 20.1 & 354.5 \\ \bottomrule
\end{tabular}
\caption{Ablation studies on different components of our proposed intention predictor.}
\label{tab:ablation}
\vspace{-8mm}
\end{table}

\noindent\textbf{Overall results.}
In Table~\ref{tab:sota}, we report the overall performance of different agents on 4,962 validation episodes across 12 different scenes. The pioneer agent achieves the best results on all success metrics and collides less with the human. 
However, this agent requires an oracle pathfinder and executes oracle actions that can be directly transferred to the planned location.
An interesting finding is that the previous best human following method~\cite{habitat3.0} struggles on this task, as the agent always follows after the human and keeps a certain distance. However, in the \sexyname task, we require the agent to arrive at the goal location actively and the distance to the target needs to be within 1 meter.

Equipped with an intention-aware human perception module, our \sexyname agent surpasses all baseline methods without any oracle input, even compared with a well-trained open-vocabulary object navigation agent~\cite{ZSON} powered by a LLM-based intended-object predictor. Besides, this agent achieves less average navigation steps, indicating that it is more intelligent at predicting where the human is intended to go based on its early observation, and acts more quickly. We attribute this to the effectiveness of our proposed intention-aware human perception module. 

\noindent\textbf{Ablation studies on the \sexyname agent.}
We provide in-depth analysis to determine the effectiveness of our \sexyname agent.
All agents are trained for 50 epochs using expert trajectories.
In Table~\ref{tab:ablation}, we report evaluation results of 5 different agent variants that remove a specific part of the intention prediction module (\ie intended object, activity, and trajectory).
There are two main findings from the results. Firstly, all types of intention representations are helpful in determining the human's destination, 
as the additional prediction results boost the agent's navigation success rate while reducing the navigation step. 
Combining all prediction results yields the best performance, indicating that predicting human intention from different perspectives based on visual observations is crucial for our \sexyname task.
We also interestingly find that the agent with more intended prediction input possesses more chances to collide with the human. This is a non-trivial trade-off since the agent has to seek a balance between walking close to the human intended trajectory and avoiding conflict.

Secondly, the intended object category and short-term trajectory matter a lot in representing human intention. As shown in Table~\ref{tab:ablation}, 
adding predicted intended activity brings a +1.8\% improvement in FASR and +1.8\% in RASR (comparison between Exp. \#3 and \#4); while adding predicted intended object brings a larger improvement (+3.1\% in FASR and +3.0\% in RASR comparing Exp. \#2 and \#4). 
We believe that determining the object category facilitates the agent to find and reach the goal position accurately. The predicted intended activity and object are complementary. Using these two predicted intentions simultaneously brings larger improvements.
Additionally, how to predict short-term humanoid future trajectory according to previous observations remains an interesting problem as it contributes a lot to the agent's performance in this task (+2.3\% improvement in FASR and +2.8\% in RASR). The point indication is a more efficient and accurate guidance for the navigation agent. 
All these findings reveal that the key challenge of the \sexyname task is how to predict a definite indication of the human's intended destination in terms of both short-term human trajectory and a long-term intended object category.
%

\noindent\textbf{Failure case analysis.}
We manually analyze the failure cases of \sexyname agent and find that the main causes can be summarized as follows: 1) Conflicts with human actions, leading to collisions, which account for failures in 20\% of the episodes. 2) The agent occasionally reaches a corner or narrow spaces that haven't been seen in the expert trajectories, making the agent stuck. This situation accounts for 30\% of the episodes 3) The agent arrives at the wrong destination and halts, which accounts for more than 50\% of the episodes. Taking into account the failure cases, there is still considerable room for worthy research on this task.

\begin{wrapfigure}{r}{6.8cm}
    \centering
    \vspace{-3mm}
    \includegraphics[width=0.75\linewidth]{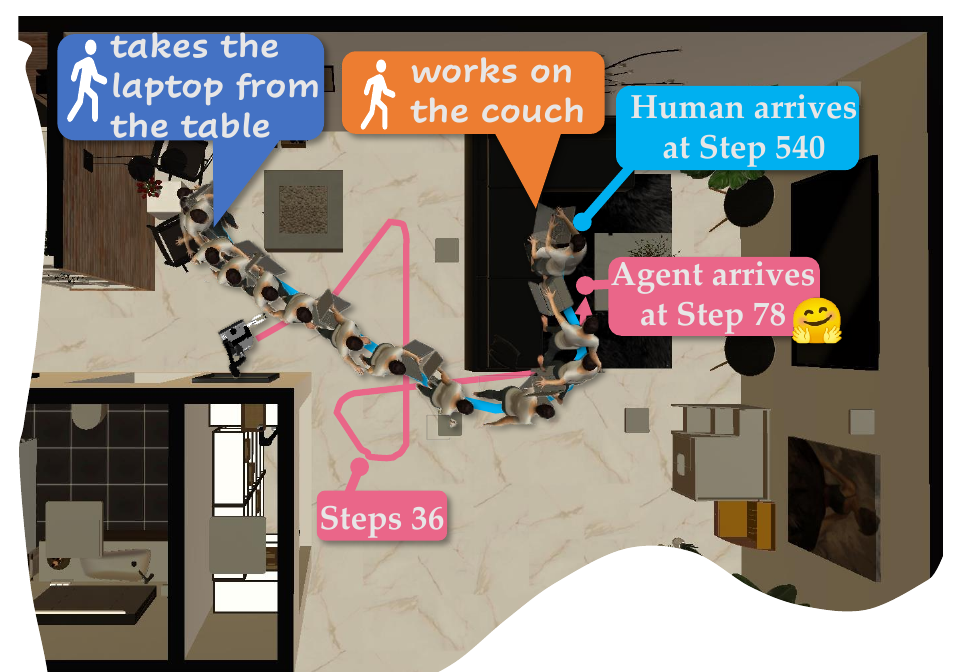}
    \caption{Visualization of a testing episode.
    }
    \label{fig:visual}
    \vspace{-6mm}
\end{wrapfigure}

\noindent\textbf{Qualitative results.}
In Figure~\ref{fig:visual}, we visualize both human and agent trajectories in a validation episode. Since the agent arrives at the destination at very low time steps, the intersections between trajectories do not lead to collision. The \sexyname agent first explores its surroundings and seeks information from the human. After the human retrieved the laptop from the table, the agent turned around and directly navigated to the table (at the 36th step). We believe that the prediction target of both long-term and short-term intention endows the agent with such ability.
\vspace{-2mm}

\section{Conclusion}
\vspace{-2mm}
In this paper, we propose a novel collaborative navigation benchmark, \sexyname, aiming to promote the development of efficient human-robot collaboration. 
Compared to previous works towards human-centered navigation, our proposed collaborative navigation task is much more challenging as it introduces diverse humanoids with realistic human behavior. 
To achieve this, we designed a novel data collection pipeline based on the success of Large Language Models and simulation platforms and constructed a dataset for this task.
Furthermore, we conduct preliminary research on different intention-unaware baselines and our proposed intention-aware agent, hoping to inspire further research for developing intelligent agents for human-robot collaboration.

\newpage
\bibliographystyle{abbrv}
{
	\small
	\bibliography{arxiv}
}

\newpage
\appendix
\onecolumn
\begin{center}
	{
		\Large{\textbf{Appendix for CoNav\\ A Benchmark for  Human-Centered Collaborative Navigation}}
	}
\end{center}
\vspace{15 pt}

This is the supplementary material for paper "CoNav: A Benchmark for Human-Centered Collaborative Navigation". We organize our supplementary material as follows:

\begin{itemize}[leftmargin=*]
    \item In Sec.~\ref{supp:data-generation}, we provide more details on the prompts for generating activity text.
    \item In Sec.~\ref{supp:dataset-analysis}, we provide more details on the dataset analysis.
    \item In Sec.~\ref{supp:humanoid-animation}, We provide more qualitative Results of Humanoid Animation.
    \item In Sec.~\ref{supp:limitation}, we discuss the limitations and future work of our work.
    \item  In Sec.~\ref{supp:broader-impacts}, we discuss the broader impacts of this work.
\end{itemize}

\section{More Details on LLM-Assisted Data Generation}
\label{supp:data-generation}
\begin{figure}[h]
    \centering
    \includegraphics[width=1\linewidth]{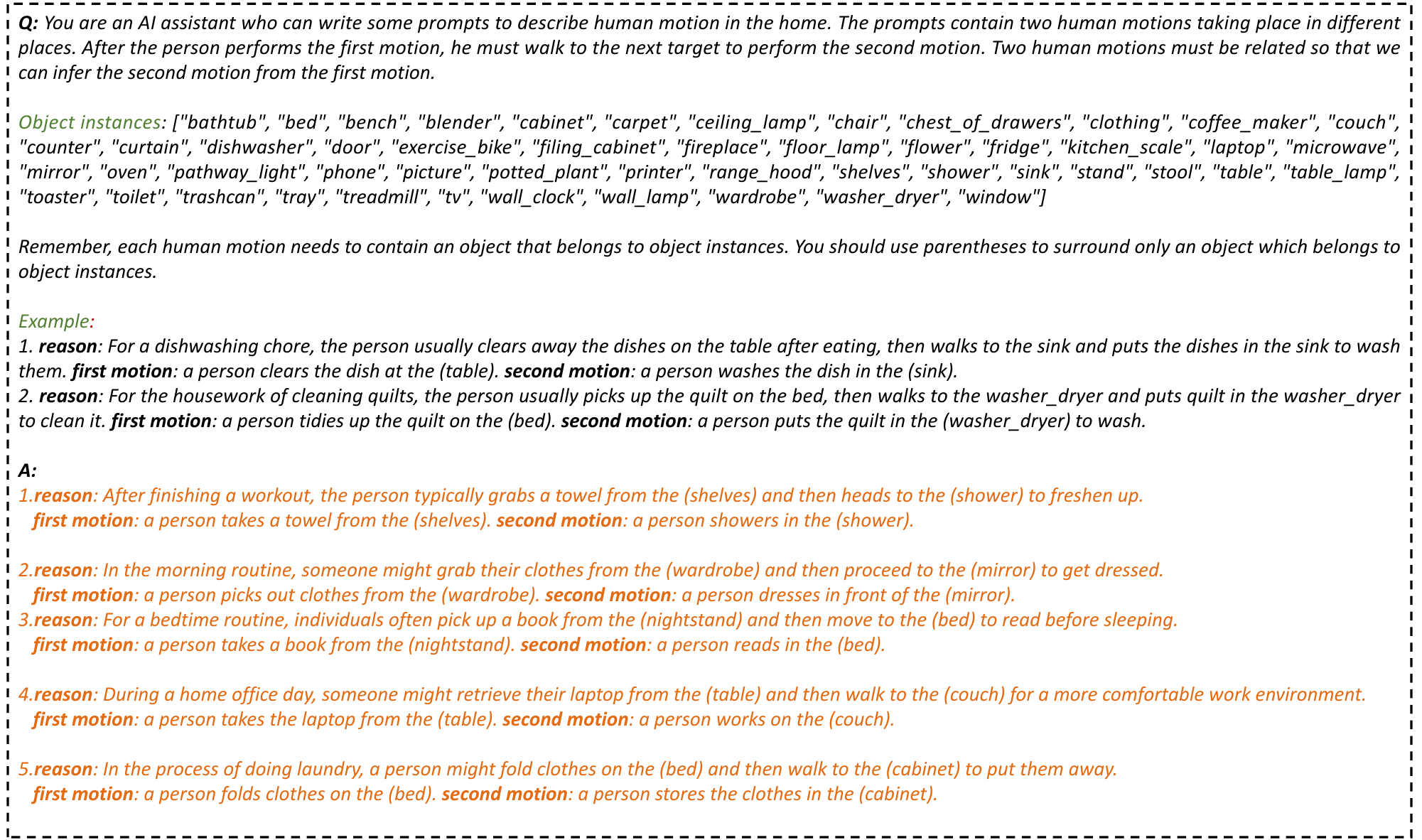}
    \caption{
    The prompt for generating the two-stage activity conditioned on the destination object set extracted from the environment and few-shot examples.
    }
    \label{fig:prompt}
\end{figure}

In Fig~\ref{fig:prompt}, we show the prompts for generating activity text and response examples of LLM. We provide few-shot examples as the context and use chain-of-thought techniques to reduce hallucinations. Next are the details of how we generate prompts. First, We ask ChatGPT as an AI assistant to help us generate text descriptions of household activities with objects. Then we ask ChatGPT to pick out the object from Object instances to avoid generating unwanted objects not in the scene. To generate two-stage activities with semantic relationships, we let ChatGPT generate the reasons why the two activities are related first so that reasonable activities can be generated based on the reasons. We also use a few-shot prompt to further limit and prompt the content generated by ChatGPT.

\begin{figure}[t]
    \centering
    \includegraphics[width=1\linewidth]{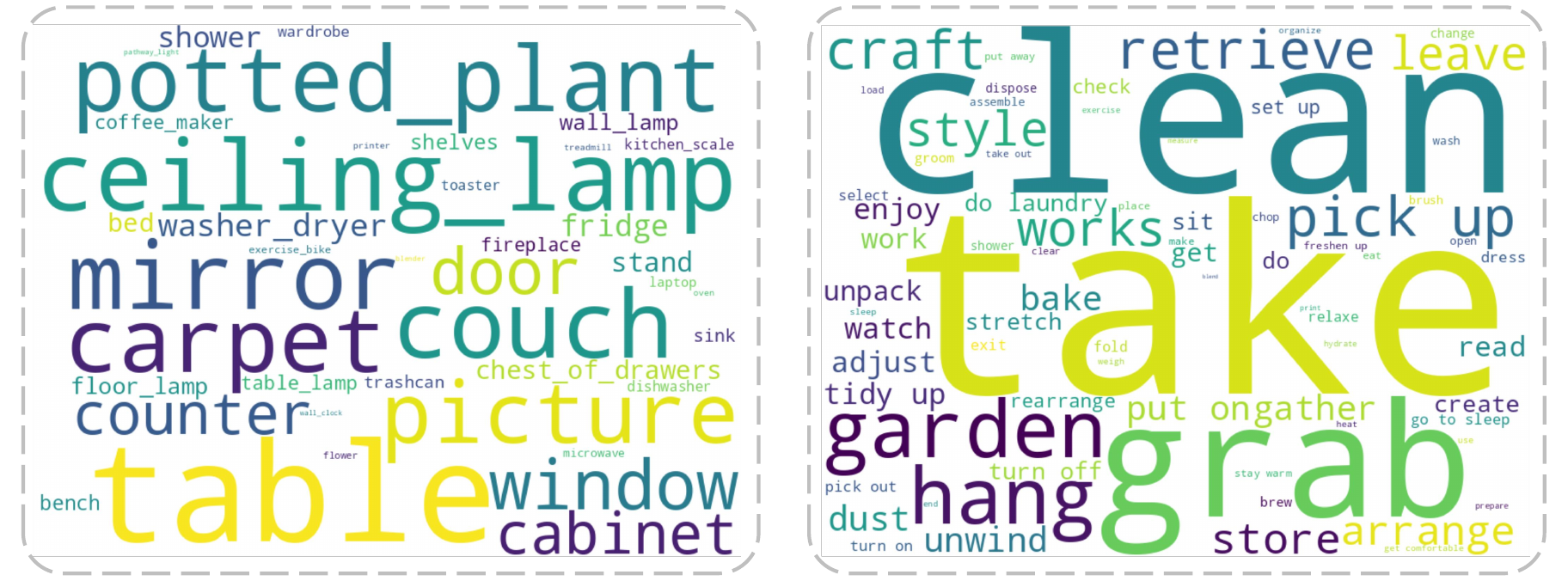}
    \caption{
    Wordcloud of intended object (left) and action (right) in the CoNav dataset.
    }
    \label{fig:wordcloud}
\end{figure}

\begin{figure}[t]
    \centering
    \includegraphics[width=1\linewidth]{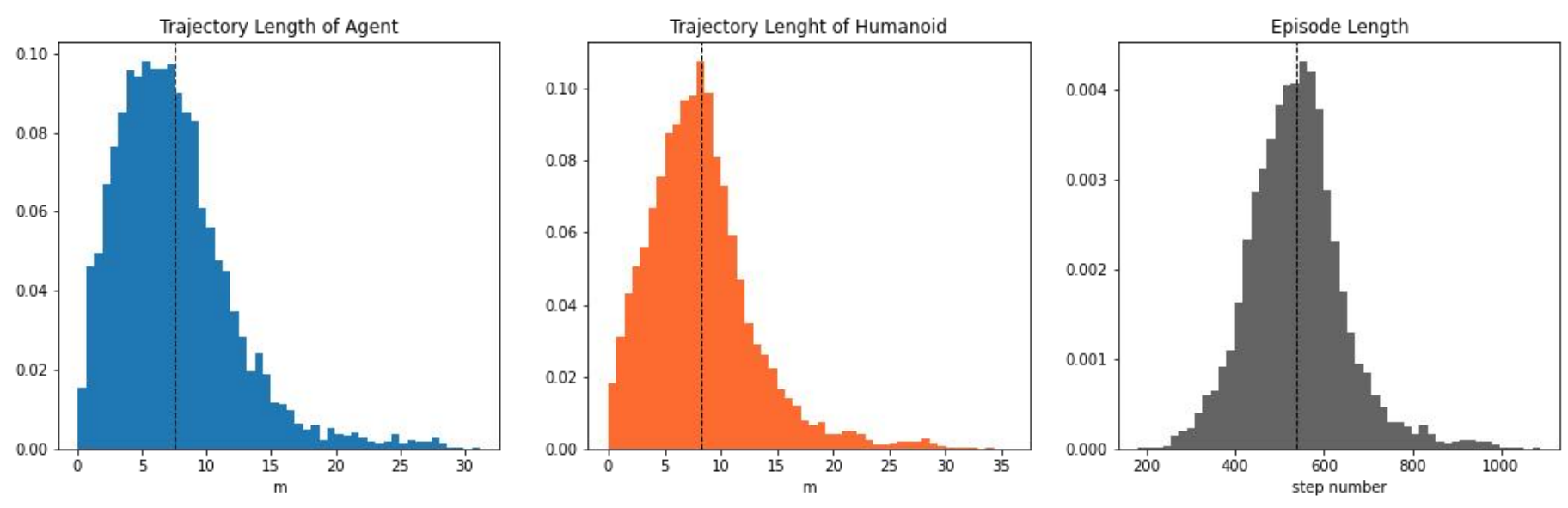}
    \caption{
    Distribution of navigation trajectory length and episode length in the CoNav dataset. \textbf{Left}: Distribution of agent navigation trajectory length. \textbf{Middle}: Distribution of humanoid navigation trajectory length. \textbf{Right}: Distribution of episode length. ``Trajectory Length" refers to the geodesic distance of the entire navigation trajectory from the starting position to the target object position. ``Episode Length" indicates the number of steps in an episode. The dashed lines represent the respective mean values.
    }
    \label{fig:distribution}
\end{figure}

\begin{figure}[t!]
    \centering
    \includegraphics[width=1\linewidth]{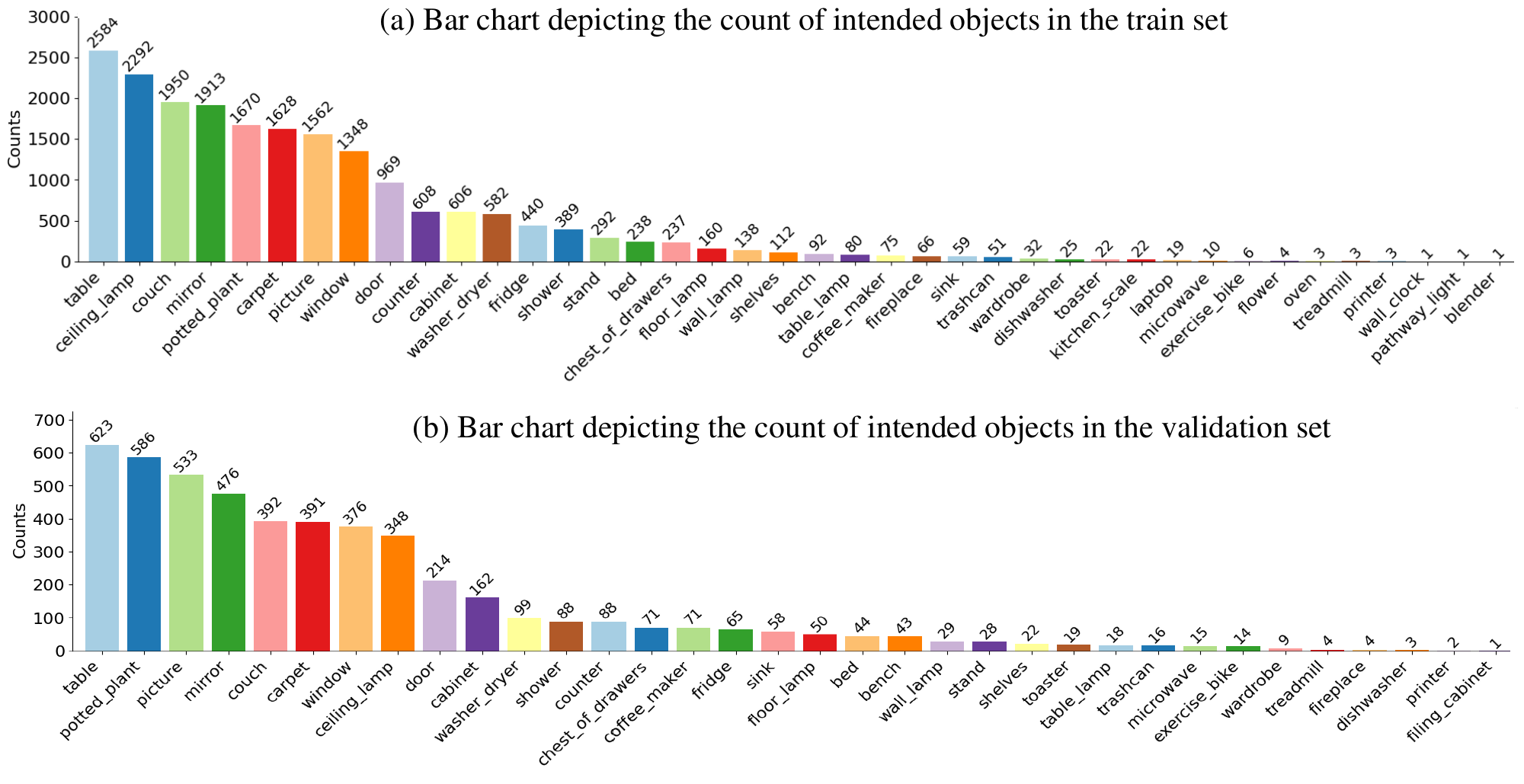}
    \caption{
    Distribution of intended objects in (a) training set and (b) evaluation set.
    }
    \vspace{1mm}
    \label{fig:object}
\end{figure}

\begin{table}[t!]
\caption{The statistics of intended activity}
\label{tab:activity}
\centering
\begin{tabular}{@{}lll@{}}
\toprule
activity                                                  & train & val \\
\toprule
a person cleans near the ceiling\_lamp                 & 2292  & 348 \\
a person gardens near the potted\_plant                & 1670  & 586 \\
a person hangs the artwork on the picture             & 1529  & 513 \\
a person cleans the carpet                            & 1377  & 350 \\
a person crafts on the table                          & 1347  & 256 \\
a person works near the window                        & 957   & 247 \\
a person styles their hair in front of the mirror     & 951   & 162 \\
a person leaves through the door                      & 808   & 168 \\
a person stores items in the cabinet                  & 478   & 145 \\
a person bakes on the counter                         & 473   & 76  \\
a person puts on the tie in front of the mirror       & 470   & 81  \\
a person unwinds on the couch                         & 469   & 151 \\
a person tidies up the cushions on the couch          & 458   & 75  \\
a person watches on the couch                         & 427   & 62  \\
a person cleans the window                            & 391   & 129 \\
a person gets cheese from the fridge                  & 331   & 40  \\
a person does laundry with the washer\_dryer           & 323   & 44  \\
a person creates on the table                         & 322   & 132 \\
a person takes a shower in the shower                 & 304   & 50  \\
a person enjoys a hot drink on the table              & 275   & 50  \\
a person unpacks groceries on the table               & 272   & 71  \\
a person does laundry in the washer\_dryer             & 259   & 55  \\
a person stretches on the carpet                      & 251   & 41  \\
a person rearranges the books in the chest\_of\_drawers & 237   & 71  \\
a person sets up on the stand                         & 230   & 22  \\
a person works on the couch                           & 217   & 44  \\
a person gathers snacks from the table                & 217   & 44  \\
a person cleans the mirror                            & 197   & 106 \\
a person goes to sleep in the bed                     & 176   & 36  \\
a person dresses in front of the mirror               & 164   & 45  \\
a person exits through the door                       & 153   & 39  \\
a person reads near the wall\_lamp                     & 138   & 29  \\
a person works on the puzzle under the floor\_lamp     & 126   & 45  \\
a person relaxes on the couch                         & 114   & 23  \\
a person reads on the couch                           & 112   & 22  \\
a person selects a movie from the shelves             & 112   & 22  \\
a person stores the clothes in the cabinet            & 102   & 12  \\
a person enjoys the snack on the couch                & 100   & 9   \\
a person takes a break at the table                   & 98    & 24  \\
a person unpacks groceries on the counter             & 95    & 5   \\
a person sits on the bench                            & 92    & 43  \\
a person turns off the table\_lamp                     & 80    & 18  \\
a person grooms in front of the mirror                & 68    & 15  \\
a person grabs a cold drink from the fridge           & 64    & 14  \\
a person showers in the shower                        & 58    & 30  \\
a person checks their appearance in the mirror        & 57    & 43  \\
a person brews coffee in the coffee\_maker             & 52    & 45  \\
a person disposes of the wrapper in the trashcan      & 51    & 16  \\
a person enjoys snacks on the couch                   & 41    & 4   \\
a person assembles ingredients from the fridge        & 40    & 7         \\ \bottomrule
\end{tabular}
\vspace{10mm}
\end{table}

\section{More Details about Dataset Analysis}
\label{supp:dataset-analysis}
In total, we collected 20,293 navigation episodes in train split with
an average length of 539 steps. As illustrated by Figure \ref{fig:wordcloud} and Figure \ref{fig:distribution}, the CoNav dataset is the most diverse and closest to our human's indoor daily activities to the best of our knowledge. We first analyze the word cloud distributions of destination objects and actions. For instance, in an episode where the motion involves ``a person picks up a book from the shelves, a person reads on the couch," the actions observed are ``pick up" and ``read" with the destination object being ``couch". Then, we investigate the distribution of navigation trajectory length for both agent and humanoid within the CoNav dataset as well as the distribution of episode length. These two figures demonstrate the diversity of our dataset.

Figure \ref{fig:wordcloud} illustrates the diversity in action type and object type. In each episode, the humanoid must navigate to various target objects and perform different actions. This adequately reflects the diversity of human behaviors in daily life and underscores the comprehensiveness of the benchmark.

Figure \ref{fig:distribution} showcases the diversity in trajectory lengths and episode lengths. These trajectories include activities with both long and short horizons, providing a more comprehensive reflection of the agent's ability to recognize human intentions and assist in task completion.

Figure \ref{fig:object} showcases the diversity of destination objects in the train set and the validation set. In the training split, there are 40 types of destination objects while the validation split features 34 types, 33 of which are overlapped.

As in Table~\ref{tab:activity}, we counted the number of occurrences of the 107 intended activities generated by LLM in the training set and evaluation set and listed the top 50 of them. As we can see, our activity descriptions are of various types and have reasonable intentions. They appear in both the training set and the test machine.

\begin{figure}[t]
    \centering
    \includegraphics[width=1\linewidth]{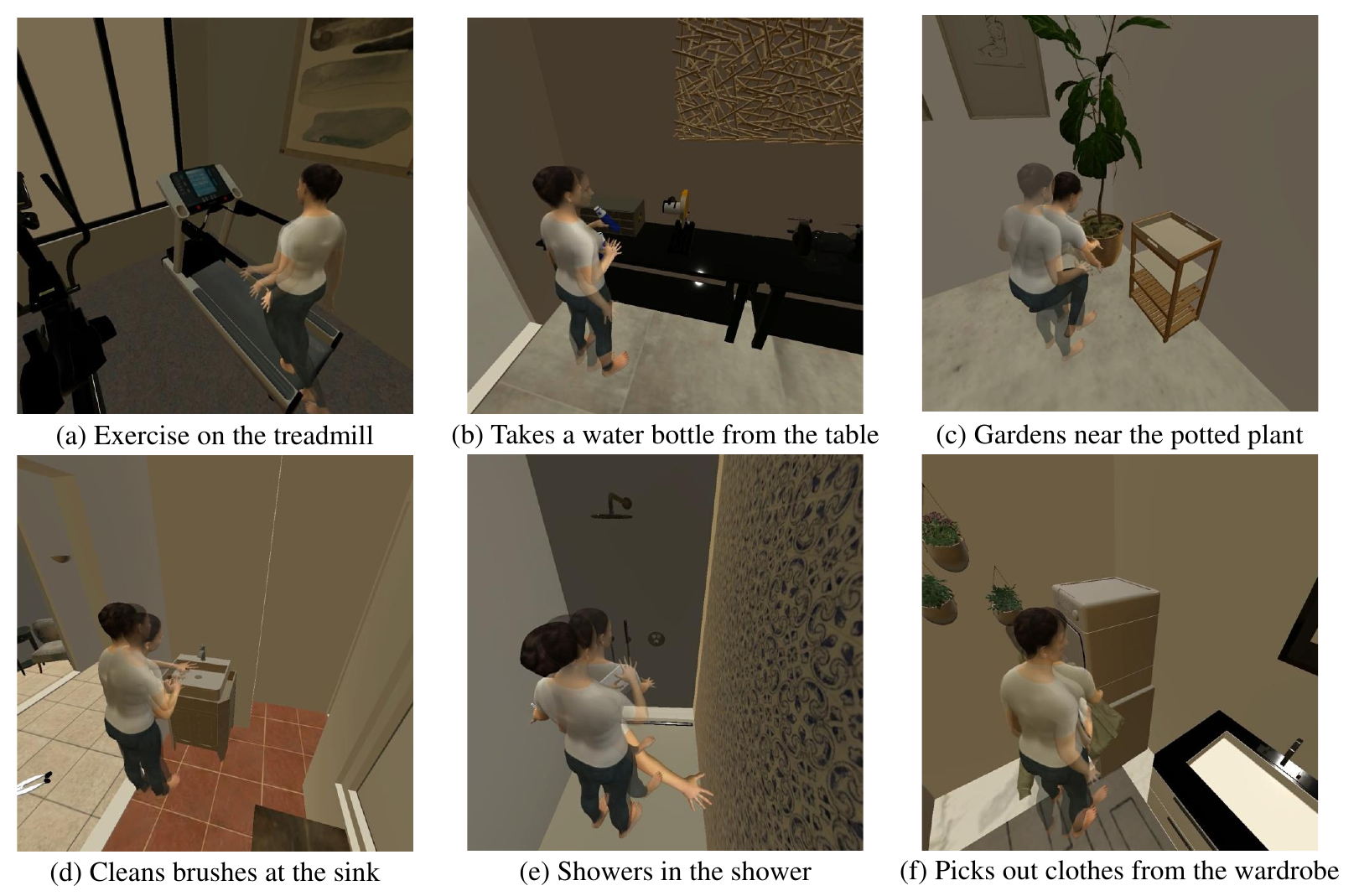}
    \caption{More visualization results of our CoNav dataset about synthesized humanoid animations.
    }
    \label{fig:action}
\end{figure}

\section{Qualitative Results of Humanoid Animation}
\label{supp:humanoid-animation}
\begin{figure}[t]
    \centering
    \includegraphics[width=1\linewidth]{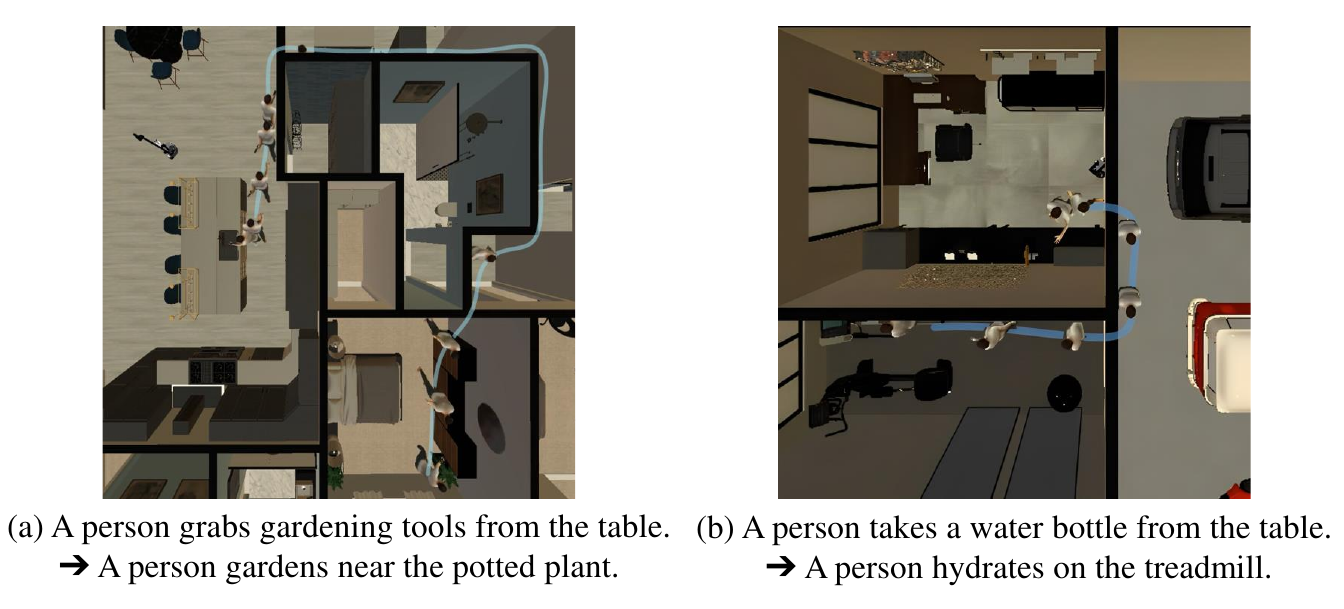}
    \caption{More visualization results of our CoNav dataset about the whole trajectory with 2 generated activities.
    }
    \label{fig:activities}
\end{figure}

We show more visual results of our Conav dataset in Figure~\ref{fig:action} and Figure~\ref{fig:activities}. 
Figure~\ref{fig:action} shows the effect of human motion generated by the text-motion generation model. Generated motions are realistic diverse, and harmonious in the simulation environment. The generated motions can match the given text which allows the agent to understand the meaning of human motions. Figure~\ref{fig:activities} shows the complete trajectory of a two-stage activity performed by a humanoid in an episode. In subgraph (a), the humanoid grabs the garden tool from the table. Then he walks along the short path trajectory 
 to cultivate the potted plant. Subpicture (b) shows a human picking up a water bottle from the table. Then walk to the treadmill to run. He picks up a water bottle because he may need to replenish water while exercising on the treadmill. The connection segment motions generated by LLM are contextual, so the agent can deduce the humanoid's next intention based on the first motion and navigate to the destination in time to help humans.

\section{Limitation and Future Work}
\label{supp:limitation}
In this paper, we focus on how to infer human intentions and navigate to target locations in time. This is the first step towards realizing human-machine collaboration, but we lack research on the interaction between robots and humans. At the same time, if there are occlusion issues in navigation, it is very difficult to predict future intentions from observed human actions, which may lead to a lower navigation success rate. Notice that in reality, humans communicate to clarify their intentions. In the future, we will expand our work to introduce dialogue information to improve the robustness and success rate of navigation.

\section{Broader Impacts}
\label{supp:broader-impacts}
The \sexyname benchmark we proposed provides a novel perspective for studying human collaborative navigation, which can reason about intentions and navigate to the destination using only human motion. However, our proposed dataset construction pipeline may be used for malicious purposes, \eg generating convincing fake actions for spoofing.




\end{document}